\documentclass{article}

\usepackage{arxiv}

\usepackage[utf8]{inputenc} 
\usepackage[T1]{fontenc}    
\usepackage{hyperref}       
\usepackage{url}            
\usepackage{booktabs}       
\usepackage{amsfonts}       
\usepackage{nicefrac}       
\usepackage{microtype}      
\usepackage{lipsum}
\usepackage{graphicx}
\graphicspath{ {./images/} }
\usepackage{amsmath}
\usepackage{array}

\usepackage[ruled, vlined]{algorithm2e}
\usepackage{color}

\usepackage{colortbl}
\newcolumntype{P}[1]{>{\centering\arraybackslash}p{#1}}
\newcolumntype{R}[1]{>{\flushrifht\arraybackslash}p{#1}}

\definecolor{c_diag}{rgb}{0.85,0.89,0.953}
\definecolor{t_diag}{rgb}{0,0,0}
\definecolor{c_ndiag}{rgb}{.95,0.95,0.95}

\definecolor{c_hlight}{rgb}{.886,0.9411,0.851}
\definecolor{c_hdark}{rgb}{.663,0.82,0.557}
\definecolor{c_rest}{rgb}{.984,0.898,0.839}

\newcommand{\ndiag}[1]
{
	\normalsize\cellcolor{c_ndiag}\textcolor{t_diag}{#1}
}

\newcommand{\hlight}[1]
{
	\normalsize\cellcolor{c_hlight}\textcolor{t_diag}{#1}
}

\title{Class--specific feature selection\\ for classification explainability}

\author{
 Jes\'us S. Aguilar--Ruiz \\
  School of Engineering\\ Pablo de Olavide University\\ 
  ES-41013 Seville, Spain \\
  \texttt{aguilar@upo.es} \\
}

\begin{document}
\maketitle
\begin{abstract}
Feature Selection techniques aim at finding a relevant subset of features that perform equally or better than the original set of features at explaining the behavior of data. Typically, features are extracted from feature ranking or subset selection techniques, and the performance is measured by classification or regression tasks. However, while selected features may not have equal importance for the task, they do have equal importance for each class. 

This work first introduces a comprehensive review of the concept of class--specific, with a focus on feature selection and classification. The fundamental idea of the class--specific concept resides in the understanding that the significance of each feature can vary from one class to another. This contrasts with the traditional class--independent approach, which evaluates the importance of attributes collectively for all classes. For example, in tumor prediction scenarios, each type of tumor (class) may be associated with a distinct subset of relevant features. These features possess significant discriminatory power, enabling the differentiation of one tumor type from others (classes). This class--specific perspective offers a more effective approach to classification tasks by recognizing and leveraging the unique characteristics of each class.

Secondly, classification schemes from one--versus--all and one--versus--each strategies are described, and a novel deep one--versus--each strategy is introduced, which offers advantages from the point of view of explainability (feature selection) and decomposability (classification). Thirdly, a novel class--specific relevance matrix is presented, from which some more sophisticated classification schemes can be derived, such as the three--layer class--specific scheme.

The potential for further advancements in this area is wide and will open new horizons for exploring novel research directions in interdisciplinary fields, particularly in complex, multiclass hyperdimensional contexts.
\end{abstract}


\section{Introduction} \label{sec:introduction}
Feature selection techniques \cite{Liu2007,Li2017} have been widely discussed in the scientific literature as a task prior to classification, where a subset of variables is extracted from the original set of variables, such that with less information (only relevant variables) similar knowledge (comparable classification performance) is maintained. In many applications, especially those involving datasets containing a high number of variables, the use of feature selection techniques is crucial. This is not only because a prediction model can often provide similar or even better results with less information, thus improving the computational efficiency of learning, but also because a smaller subset of variables can enhance our understanding of the data's behavior,  i.e., provide us with greater explainability \cite{Barredo2020}. 


The genomics boom has led to the emergence of datasets with thousands of variables, exacerbating the well--known issue of ``the curse of dimensionality''. Furthermore, transposable elements have further increased dimensionality to the order of millions of variables \cite{Wellinger2022}. This new scenario has introduced fresh challenges for existing algorithms and has spurred research into the development of efficient techniques capable of handling such a vast number of features, from two perspectives: performance and explainability. Selected variables might be unequally important for the classification task (e.g. when they are selected from a feature ranking, and each variable has an associated weight or quality measure). However, the selected variables are assumed to be equally important for each class, independently, which is a strong assumption. To further illustrate this, envision a medical diagnosis situation involving various diseases. Feature selection techniques would identify a set of features that are effective for diagnosing all these diseases.  However, certain attributes might be significant indicators for a rare disease, while being less relevant for common ailments. It would then be appropriate to prioritize attributes that are excellent at distinguishing a rare but critical disease, even if they are less effective in distinguishing other, more common conditions. In a severe case such as tumor prediction, different tumors might correlate with distinct subsets of relevant variables that are particularly effective in distinguishing them from other tumors. In summary, while feature selection techniques are adept at finding variables relevant for classification on a global scale, they do not necessarily pinpoint which variables are pivotal for each class on a local level.


In this context, we focus on identifying variables that are relevant to each class independently, a process known as \textit{class-specific feature selection}. This concept is crucial in situations where the goal is to determine the unique discriminatory power of attributes for specific classes within a multi--class problem. Traditional attribute selection methods typically assess the overall discriminatory power across all classes. However, this approach may not be suitable when certain attributes are particularly relevant for distinguishing one class from others. Class--specific feature selection evaluates and selects attributes based on their ability to distinguish a particular class of interest, even if these attributes are less important for other classes. This method is especially valuable in fields like healthcare, where the significance of different attributes can vary greatly in identifying specific medical conditions or diseases.


Class--specific attribute selection is particularly useful for prioritizing attributes that are essential for detecting rare diseases, even if they are less informative for other diagnoses. This approach empowers the making of more targeted and context--specific decisions during the feature selection process. In the field of machine learning, particularly in classification problems involving multiple classes, the selection of appropriate features is often critical to enhancing both the performance and the interpretability of the models. Class--specific feature selection presents a nuanced approach to model development, offering several key advantages over traditional feature selection methods:

\begin{itemize}
\item Enhanced Model Performance: This method can significantly improve the performance of models by ensuring they excel in classes where sensitivity is crucial. By focusing on the most relevant features for each class, it optimizes the model's ability to make accurate predictions in critical areas.

\item Improved Explainability: Class--specific feature selection also contributes to more explainable models. By clearly identifying which attributes are important for specific classes, it simplifies the task of explaining why a model makes particular predictions. This transparency is especially valuable in fields where understanding the decision--making process of a model is as important as the accuracy of its predictions.
\end{itemize}

In practical applications, class--specific feature selection can be implemented using a range of techniques and metrics, such as class--specific feature importance scores. These scores are instrumental in identifying attributes that play a significant role in the classification of a specific class, while also taking into account their relevance to other classes. This approach can lead to more effective and tailored machine learning models, especially in domains where certain classes or outcomes are of paramount importance and can guide decisions on which attributes to include or exclude for different classes.

The rest of the document is organized as follows: Section 2 briefly describes feature selection, elaborates on the class--specific concept and reviews the related work; Section 3 introduces the class--specific approaches; Section 4 discusses the relationship between class--specific feature selection and classification, and introduces novel concepts in the field; finally, the main conclusions and future work are described in the last section.


\section{Class--specific Feature Selection}

\subsection{Feature Selection}

The concept of the \textit{curse of dimensionality} originates from the foundational work of mathematician Richard Bellman in approximation theory \cite{Bellman1961}. It represents the important challenge of uncovering latent structures in variable--rich data sets. As the number of explanatory variables increases, so does the complexity of identifying these structures, closely linked to the complex task of feature selection for model fitting. Bellman's articulation of the curse of dimensionality encapsulates this exponential increase in complexity, notably apparent in complex problems harboring a multitude of variables, making dimensionality overwhelmingly unmanageable.

The early stages of machine learning were dominated by the tendency to encompass all variables, which was far from the principle of feature parsimony, crucial for predictive accuracy. Nowadays, where very high dimensional contexts exist \cite{Feltes2019}, restricting the number of input features helps mitigate overfitting and enhances the models' ability to make accurate estimates. However, although there have been many efforts to reduce the dimensionality faced by a classifier, few instead have attempted to build the classifier based on a non--homogeneous set of features, i.e., feature sets selected separately for each class, as opposed to the universal approach that performs the selection jointly for the entire dataset. 


In general, there are two main procedures to obtain a subset of relevant variables: a) Evaluate possible subsets of any size extracted from the original set of variables (preferably, all subsets) and choose the best one; b) Generate a variable ranking and select the final subset considering an incremental heuristic over the ordered list.


Generating all possible rankings of variables in a dataset is impractical due to the prohibitively high cost, which is factorial in nature, specifically $m!$, where $m$ is the number of variables. Similarly, evaluating every possible subset of variables is also infeasible, as this requires exponential computational resources, quantified as $2^m$. A more efficient approach is to independently assess each variable according to a specific evaluation criterion. This method incurs a cost of $m$ times the cost of evaluating a single variable, which typically does not exceed $n^2$, with $n$ representing the number of examples in the dataset. For instance, if the information gain criterion is applied to each variable, the algorithmic complexity would be linear with respect to both the number of variables and examples, denoted as $O(mn)$.


The challenge of variable independence assumptions in the generation of variable rankings can be addressed through a technique known as \textit{forward selection}. This approach involves the successive addition of variables that rank highly. However, this method is not without its risks. For example, the top two variables in the ranking might be correlated, leading to their selection for the final subset despite this redundancy. To enhance the effectiveness of this process, incremental heuristics can be employed. These heuristics carefully evaluate the benefits of including or excluding a higher-ranked variable. Such an approach has been shown to substantially improve outcomes \cite{Ruiz2006}.


The class--specific methodology can be integrated into two distinct procedures: subset generation and evaluation, as well as ranking generation and subset extraction. This approach focuses on the conditional selection of attributes, aiming to identify which attributes are significant for each class. This process can be conducted with or without attribute overlap and involves either the generation of rankings or the extraction of subsets. The essence of class-specific feature selection lies in its ability to discern the relevance of specific attributes in the context of each class, thereby tailoring the selection process to the unique characteristics of each class.


Feature selection techniques can be broadly categorized into two groups: class-independent and class-specific. Class--independent techniques encompass the majority of traditional feature selection methods, such as Information Gain  \cite{Quinlan1986}, $\chi^2$ Test \cite{Plackett1983}, and ReliefF \cite{Kononenko1997}, which are typically referred to as classification--specific. These techniques aim to identify a global set of attributes relevant for the entire dataset without differentiating between its classes, and they are primarily used in classification tasks.

In contrast, class--specific feature selection focuses on each class independently to ascertain a relevant subset of attributes for each class. This approach allows for a tailored analysis of each class, identifying attributes that are particularly significant for each one. The results from this process can either be utilized separately for each class or combined using an aggregation method to form a comprehensive attribute subset that encompasses insights from all classes. This distinction allows for a more tailored approach to feature selection, catering to the specific needs of different types of classification problems.

In the context of binary datasets with two class labels, a variable that is relevant for one class, demonstrating discriminatory power, is inherently relevant for the other class as well. However, the scenario becomes more complex in multi--class problems, where datasets feature more than two class labels. In these cases, a variable might be highly relevant for one class but not for others. Therefore, it is crucial that the variables selected for such datasets collectively possess the ability to discriminate each class effectively.

\subsection{Related work}\label{subsec:related_work}


Baggenstoss made significant contributions to the development of a class-specific model of behavior, focusing on the separation of attributes by class. In his earlier works \cite{Baggenstoss1998, Baggenstoss1999}, he introduced the concept of  \textit{class--specific} modeling, aimed at estimating low--dimensional probability density functions while maintaining the theoretical effectiveness of classification. Subsequent research by Baggenstoss \cite{Baggenstoss2003,Tang2016,Baggenstoss2022} further elaborated the class--specific approach by employing an invertible and differentiable multidimensional transformation to generate new features. This approach represents a feature extraction procedure rather than a mere selection of existing attributes from the dataset. In this context, the generated features are class--specifically derived or extracted, signifying that they are not subsets of the original dataset attributes, but are instead new features created through transformation processes.



The method introduced by Oh et al.  in\cite{Oh1998,Il-Seok1999} for handwriting recognition, named \textit{class--dependent} (unlike the \textit{class--common} approach), takes for each class $c_i$ the best features that provide large class separation between $c_i$ and the class group $\{c_j \mid 1\leq j \leq L, j\neq i\}$, and then builds $L$ neural networks with two possible outputs (positive and negative class, respectively), whose results will be further combined by selecting the most frequent output class (i.e., an ensemble strategy). To calculate the separation between two classes, the unit step kernel function is used to estimate the probability distribution included in the class separation mathematical expression. While the method was designed to generate multiple classifiers based on multilayer perceptrons, Fu and Wang \cite{Fu2002} later adopted the same idea to build a single radial basis function neural network. Sali and Ullman \cite{Sali1999} proposed the identification of common substructures, called \textit{fragments}, that are specific to a class of objects in object classification. Two--class classifiers were used for the classification of four glioma types by selecting one, two and three features (genes) while applying a one--against--all strategy \cite{Kim2002} (see below).

A general framework for class--specific feature selection, based on the one--against--all strategy is proposed in \cite{Pineda2011}. They transform the $L$--class problem into $L$ binary problems, one for each class, and then the classification is performed by an ensemble strategy. For each class, after one--against--all class binarization, an oversampling technique is applied to balance the dataset, and then a feature selection method extracts the relevant features for that class. 

When the number of features is very high, the choice of the number of features selected for each class is critical. Several experiments with very high dimensional datasets (three genomics datasets with more than two thousands genes) were conducted in \cite{Roy2013}. The authors used a separability index to measure the goodness of a feature for a class, although the method finally selected the subset of features following a sequential forward strategy that stopped when a threshold was met. Nine genomics datasets were used by Nardone et al. \cite{Nardone2019} in their Sparse--Modeling Based Approach for Class Specific Feature Selection (SMBA--CSFS), where the most representative features for each class--sample set of the training set are retrieved without taking into account other classes (no binarization). However, the number of features selected ranged from 1 to 80, presenting classification performance results for the top 20 features using different feature selection techniques because the approach exhibited good performance with that number of features. A one--against--all strategy was also used by Yuan et al. \cite{Yuan2020} to tackle with cancer detection from gene expression profiles by means of embedding elastic net in probabilistic support vector machines. Although the elastic net reduces the number of attributes for each class (after dataset binarization), the choice of initial parameters (in particular, the amount of shrinkage and the balance between the L1 and L2 penalty terms) is critical.

Recently, Ma et al. \cite{Ma2023} introduced a class--specific feature selection method by using some information--theoretic measures with respect to a particular class in order to choose a well--suited feature subset for each class (a sequential forward strategy to select top--ranked feature for each class was used). The work shows examples in which there is a difference between the feature subset selected for all classes (class--independent) and the different feature subsets for each class (class--specific), that suggests to focus on class--specific feature relevance. Then, $L$ classifiers are used (one for each class) and the predictions are combined by an aggregation method. 

\begin{figure}[t]
\centering
\includegraphics[width=0.7\linewidth]{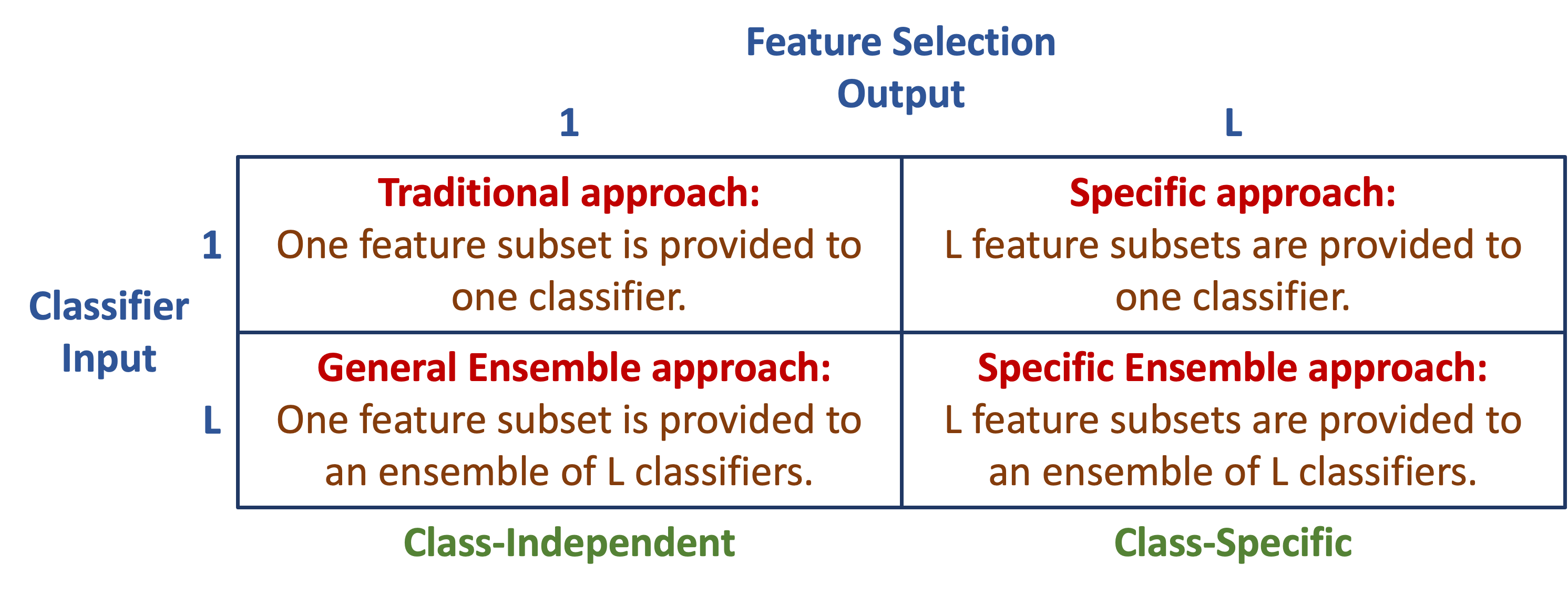}
\caption{The taxonomy of approaches based on the number of feature subsets can be categorized into two primary types: 1  (the same set of features is employed universally across all L classes) or L (one for each L classes).}
\label{fig:taxonomy}
\end{figure}

In summarizing the limited scientific literature on this topic, it is possible to categorize various methods based on the number of attribute subsets involved, applicable to both the feature selection technique and the subsequent classification technique. This classification is illustrated in Figure \ref{fig:taxonomy}. Intriguingly, all the techniques reviewed in the analysis fall into the (L, L) category. This denotes a class--specific ensemble approach, where multiple classifiers handle multiple subsets of attributes, with each subset being class--specific.

Conversely, none of the techniques reviewed fit into the (1, L) category. For a technique to qualify as (1, L), it must involve a single classifier that is capable of utilizing distinct subsets of attributes tailored to each class. The absence of techniques in the (1, L) category suggests a potential area for future research, exploring the feasibility and effectiveness of employing a single classifier to manage class--specific attribute subsets in classification tasks. This observation highlights an interesting gap in the current landscape of feature selection and classification methods.

This taxonomy helps in understanding and categorizing feature selection approaches based on their strategy towards handling class--specific information, ranging from a universal approach (using the same feature set for all classes) to a more individualized approach (each class has its own dedicated set of features).

\section{Methods}

\subsection{Notation}

Let $D=(E,F,\upsilon,\omega)$ be a dataset, where $E$ is the set of example (or instance) identifiers, $F$ is the set of feature (variable) identifiers, $\upsilon:E\times F \rightarrow \mathbb{R}$ is the function that assigns a real value to a pair $(e,f)$, where $e\in E$ and $f\in F$. Within the field of supervised learning, when $\omega:E \rightarrow \mathbb{L}$ assigns a class label $c$ to an example $e$, with $\mathbb{L}=\{c_1,\dots,c_L\}$, then it is a classification problem; otherwise, if $\mathbb{L}\subseteq\mathbb{R}$, then it is a regression problem. In absence of example identifiers (e.g. patient identifiers) or feature identifiers (e.g. gene names), it is convenient to use $E=\{e_1,\dots,e_n\}$ and $F=\{f_1,\dots,f_m\}$, respectively. Henceforth, $\upsilon(e_i,f_j)$ will represent a real value, and $\omega(e_i)$ will represent a class label (e.g., type of tumor). 

Both the function $\upsilon$ and the function $\omega$ are expressed in tabular form, so that $\upsilon$ is determined by a $n\times m$ matrix of real values (where $n$ is the number of examples and $m$ the number of features), and $\omega$ as a vector of $n$ values from $\mathbb{L}$. In order to simplify the notation, a function is defined to extract the values of sample $e_i$: $values: E \rightarrow \mathbb{R}^m$, such that $values(e_i)=(\upsilon(e_i,f_1),\dots,\upsilon(e_i,f_m))$.

The main goal of any classification problem is to find a general function $\Omega: \mathbb{R}^m \rightarrow \mathbb{L}$, learned from $D$, such that $\Omega(values(e_i)) = \omega(e_i)$, $\forall e_i$, where $i=\{1,\dots,n\}$, or at least it maximizes the frequency of the equality (a popular --and biased-- measure of quality of $\Omega$ is the relative frequency of success, named \textit{accuracy}).

\subsection{Class--specific approaches}

Let us assume the dataset $D$ contains $L=|\mathbb{L}|$ classes, i.e., it is a multiclass problem. Let $D_c=\left\{e\in E \mid w(e)=c \right\}$ be the subset of examples in $D$ that belong to class $c \in \mathbb{L}$. Two main algorithmic approaches can be used to address the problem of class--specific feature selection for multiclass datasets, as described next.

\subsection*{One--versus--All (OvA)}

The method transforms a $L$--class problem into $L$ binary problems in order to  determine the most relevant attributes for each class versus the others. The algorithm provides the same class to all the examples in $D \setminus D_p$ (i.e., negative examples) while maintaining the class of $D_p$ (i.e., positive examples) during binarization process ($binarize$), and then calculate the $measure$ for each partition of data (class $p$).

\begin{algorithm}
\KwData{$D$: dataset}
\KwResult{$\lambda$: metric}
\caption{One--versus--All (OvA)}
\label{alg:ova}
$\Lambda \leftarrow \emptyset$\\
\For{$j \in F$}
	{
	$\Lambda_j \leftarrow \emptyset$\\
	\For{$p\in \mathbb{L}$}
		{
		$D' \leftarrow$ binarize$(D_p, D \setminus D_p)$ \\
		$\lambda_p \leftarrow$ measure$(D', j)$ \\
		$\Lambda_j \leftarrow \Lambda_j \oplus (p, \lambda_p)$ \\
		}
	$\Lambda \leftarrow \Lambda \oplus (j, \Lambda_j$)\\
	}
\end{algorithm}

The overall complexity of the procedure shown in Algorithm \ref{alg:ova} is $O(mnL\alpha)$. It is simple to calculate the cost $C$: there will be $m$ iterations (outer loop), one for each feature, and $L$ iterations (inner loop), one for each class. Then, the function $binarize$ for $n$ examples, and the calculation of applying the $measure$ to feature $j$ with all the $n$ examples ($n \alpha$). Therefore, the overall cost is $C=mL(n+n \alpha)$.

As the number of classes increases, it is quite likely that after using OvA binarization, a progressive imbalance is also generated \cite{Japkowicz2002}, which might be a very serious problem when the number of classes $L$ is high. Furthermore, after selecting features for a two--class problem, the classifier must be trained using the original $L$ classes, i.e., a multiclass problem. However, the classifier is only robust to the reference class, because the features were selected to discriminate this class from the others.
 
\subsection*{One--versus--Each (OvE)}


This approach deals with one class against each other (pairwise), and was introduced in \cite{Il-Seok1999}. The authors defined a measurement of how well a class is separated by a feature vector from the remaining classes, by adding the separation between a class and each other.  

\begin{algorithm}
\KwData{$D$: dataset}
\KwResult{$\lambda$: metric}
\caption{One--versus--Each (OvE)}
\label{alg:ove}
$\Lambda \leftarrow \emptyset$\\
\For{$j \in F$}
	{
	$\Lambda_j \leftarrow \emptyset$\\
	\For{$p\in L$}
		{
		$\Psi_p \leftarrow \emptyset$\\
		\For{$q\in L \mid q\neq p$}
			{
    			$\lambda_q \leftarrow$ measure$(D_p \cup D_q, j)$ \\
			$\Psi_p \leftarrow \Psi_p \oplus (q, \lambda_q)$ \\
			}
		$\lambda_p \leftarrow$ aggregate$(\Psi_p)$   \hspace{3cm} $(\star)$ \\
		$\Lambda_j \leftarrow \Lambda_j \oplus (p, \lambda_p)$ \hspace{3.2cm} \\ 
		}
	$\Lambda \leftarrow \Lambda \oplus (j, \Lambda_j)$\\
	}
\end{algorithm}

Let us suppose that for $L$ classes, we have $n=n_1+\dots+n_L$, where $n_i$ is the number of examples of class $i$. The algorithm computes pairs of classes, then the overall cost $C$ with respect to the number of examples (the two inner loops, for $p$ and $q$) would be as follows:

\begin{equation}
C=\left((L-1)n_1+\sum_{i=2}^L n_i\right)+\left((L-2)n_2+\sum_{i=3}^L n_i\right)+\dots+\left((n_{L-1}+n_L)\right)=(L-1)\sum_{i=1}^L n_i=(L-1)n
\end{equation}

Counting the outer loop (for $j$) and adding the cost of the function $measure$, the overall cost is $C=m(L-1)n\alpha$, i.e., the complexity of the procedure shown in Algorithm \ref{alg:ove} is $O(mnL\alpha)$, the same as for the one--against--all procedure. The function $measure$ is applied to rows $e \in D_p \cup D_q$ and column $j$ (the attribute being analyzed) at each iteration. For instance, $measure$ might be an entropy--based function \cite{Shannon1948}. The function $aggregate$ takes for the class $p$ the measures calculated for each pair $(p,q)$ in $\Psi_p$ and provides a value of merit for class $p$. For instance, if the aggregation is done by taking the average, this results in a value that represents the average discriminating power of feature $j$ for class $p$ against all other classes $q$. 



\section{Discussion}



The most widely adopted approach in class--specific contexts is the OvA technique. From a classification standpoint, there is a substantial difference between opting for an OvA strategy versus an OvE strategy. Although both strategies yield the same type of output ---a ranking of features where each feature is assigned a quality measure--- the information encapsulated within these results differs. Algorithm \ref{alg:ova} calculates feature scores based on their ability to discriminate one class against all others collectively, while Algorithm \ref{alg:ove} determines how well each feature distinguishes one class from each of the others individually, and then the function $aggregate$ compiles these individual scores into a single value.

To illustrate how these strategies influence classification, consider a simple example: a dataset $D$ comprising four classes ($\mathbb{L}=\{A, B, C, D\}$) and nine attributes ($F=\{f_1, \dots, f_9\}$). Without loss of generality, let us suppose that all values returned by the function $measure$ fall within the range $[0,1]$. The higher the value, the greater the feature relevance. Henceforth, for simplicity, an attribute is considered relevant if it exceeds the threshold of 0.5 (values are shown in red in the next tables).

The output of a traditional feature selection technique would be as depicted in Table \ref{tab:ex_ranking_traditional}, in which only two features are relevant ($\{f_2, f_5\}$) for the dataset $D$, and for all the classes. Many feature selection techniques can provide similar results in structure, but they do not show how relevant is each feature for each class.
\begin{table}[h]
\centering
\caption{Ranking of features from a traditional feature selection technique.}
\label{tab:ex_ranking_traditional}
\begin{tabular}{|c|c|c|c|c|c|c|c|c|c|}
\hline
   $f_1$ 	& $f_2$ 	& $f_3$ 	& $f_4$ 	& $f_5$ 	& $f_6$ 	& $f_7$ 	& $f_8$ 	& $f_9$ 	\\ \hline \hline
 0.4 	& {\color{red}0.8} 	& 0.2		& 0.3 	& {\color{red}0.7} 	& 0.3		& 0.2 			& 0.3		& 0.3 	\\ \hline

\end{tabular}
\end{table}

\begin{figure}[h]
\centering
\includegraphics[width=0.35\linewidth]{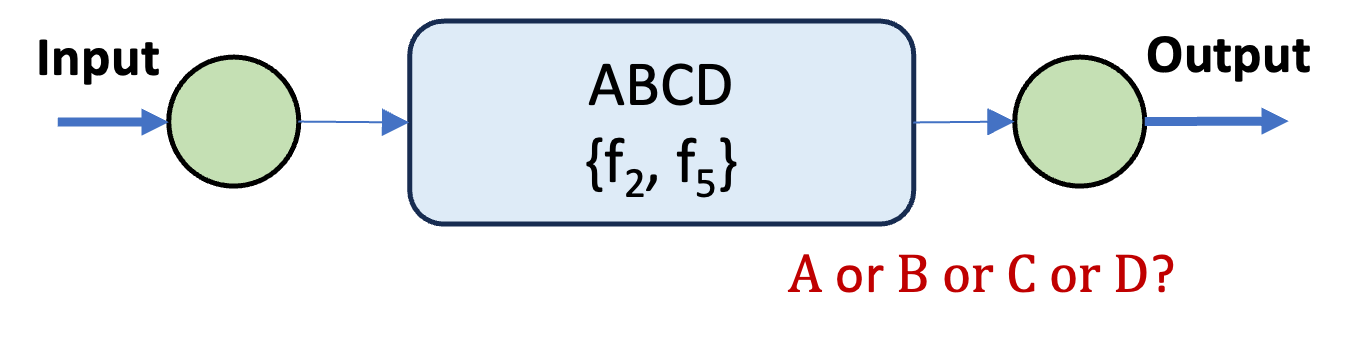}
\caption{Classification scheme for a traditional feature selection method, from the results in Table \ref{tab:ex_ranking_traditional}.}
\label{fig:ex_classifier_traditional}
\end{figure}

Figure \ref{fig:ex_classifier_traditional} shows the classification scheme, in which from the nine input features the classifier would only use the two most relevant ones to provide the output class.

Regardless of the chosen class--specific strategy, the outcome will be $\Lambda$, depicted as an $L\times m$ matrix (classes $\times$ features) of real values, as shown in Table \ref{tab:ex_ranking}. For each class the following results are obtained: \mbox{$A \rightarrow \{f_1, f_2, f_5, f_9\}$}, \mbox{$B \rightarrow \{f_2, f_5, f_7\}$}, \mbox{$C \rightarrow \{f_2, f_5\}$}, and $D \rightarrow \{f_2, f_5\}$. Only features $\{f_2, f_5\}$ are common to the four classes.

\begin{table}[h]
\centering
\caption{Example of result of class--specific strategy.}
\label{tab:ex_ranking}
\begin{tabular}{|c||c|c|c|c|c|c|c|c|c|}
\hline
   & $f_1$ 	& $f_2$ 	& $f_3$ 	& $f_4$ 	& $f_5$ 	& $f_6$ 	& $f_7$ 	& $f_8$ 	& $f_9$ 	\\ \hline \hline
A & {\color{red}0.7} 	& {\color{red}0.6} 	& 0.2		& 0.3 	& {\color{red}0.6} 	& 0.3		& 0.2 			& 0.4		& {\color{red}0.8}  	\\ \hline
B & 0.4 			& {\color{red}0.8} 	& 0.2 	& 0.4 	& {\color{red}0.7} 	& 0.2 	& {\color{red}0.6} 	& 0.2 	& 0.4				\\ \hline
C & 0.4 			& {\color{red}0.6} 	& 0.2 	& 0.5 	& {\color{red}0.7} 	& 0.5		& 0.4 			& 0.4 	& 0.4 			\\ \hline
D & 0.3 			& {\color{red}0.8} 	& 0.2 	& 0.2 	& {\color{red}0.8} 	& 0.4 	& 0.4 			& 0.2 	& 0.4 			\\ \hline
\end{tabular}
\end{table}


It can be observed in Table \ref{tab:ex_ranking} that some features are only relevant to one class (e.g., $f_7$ for class B), so they could go unnoticed in a traditional feature selection technique, which would probably only identify the features $f_2$ and $f_5$ (relevant for all classes). 



\begin{figure}[h]
\centering
\includegraphics[width=0.5\linewidth]{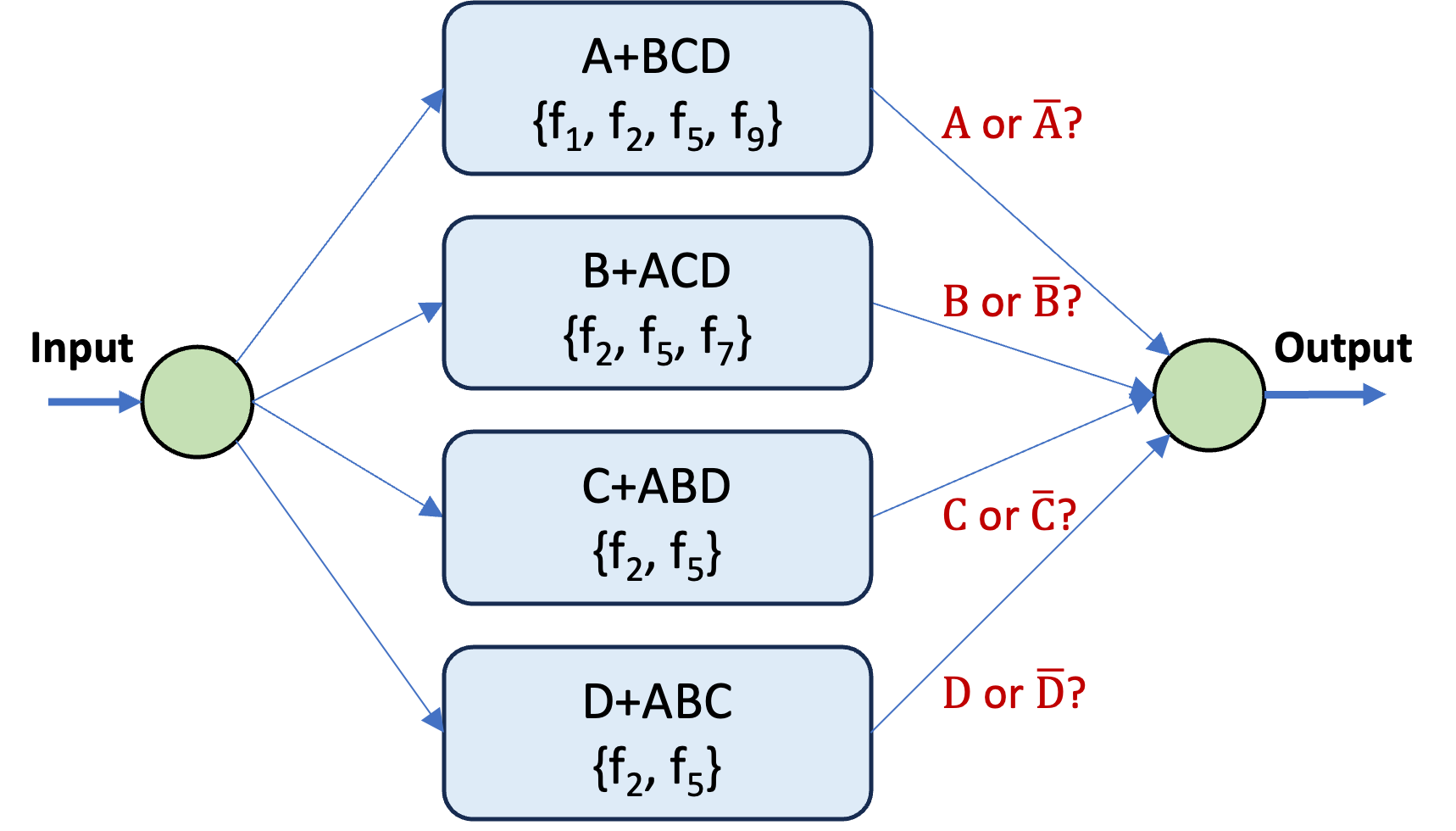}
\caption{Classification scheme from the results shown in Table \ref{tab:ex_ranking}.}
\label{fig:ex_classifier}
\end{figure}

The design of a classifier ensemble from the class--specific results will include the feature selection for each class, in such a way that each independent classifier should discriminate between one class and the rest considering only the selected features. Information contained in Table  \ref{tab:ex_ranking} can be directly transferred to the classifier ensemble depicted in Figure \ref{fig:ex_classifier}. The identifier A+BCD indicates that the classifier discriminates class A from the set of classes \{B, C, D\}, so it is only able to discriminate between A and the complement of A. Also, the relevant features for each class are included in the light blue box. At the output, an aggregation criterion should be applied to unify the result from the four partial classification outputs.

The classification scheme based on class--specific strategies not only considers the outcomes of the traditional approach, which typically highlights attributes such as $\{f_2, f_5\}$, but also underscores the significance of other attributes, namely $\{f_1, f_7, f_9\}$, for distinguishing between specific classes. This method represents a progressive step towards enhancing explainability. It can address challenges related to high dimensionality and class imbalance by adopting a more targeted approach. In line with further exploring the effects of incorporating new class--specific strategies into classification frameworks, some possible developments will be shown below.

\subsection{Deep one--against--each}

A further step in class--specific selection strategies is to explore pairwise selection in more detail. In the example with four classes, $\Psi_p$ would contain 3 pairs (see in Algorithm  \ref{alg:ove} the line marked with $\star$), and the function $aggregate$ would calculate one value from those three. The OvE strategy (Algorithm \ref{alg:ove}) does not return the discriminating power of a feature for a class against each other, but it computes for a class $p$ all the measures for the pair $(p,q)$, for all $q$, and then an aggregation function returns a single value of merit for class $p$. However, if we keep and analyze those values, we could design a better classifier ensemble, because each classifier would be able to discriminate one class from another (instead of all the others). Then, the set of pairs  $\Psi_p$ must be included in $\Lambda$, as it is shown in Algorithm \ref{alg:oves}.

\begin{algorithm}
\KwData{$D$: dataset}
\KwResult{$\lambda$: metric}
\caption{Deep one--versus--each (DOvE)}
\label{alg:oves}
$\Lambda \leftarrow \emptyset$\\
\For{$j \in F$}
	{
	$\Lambda_j \leftarrow \emptyset$\\
	\For{$p\in L$}
		{
		$\Psi_p \leftarrow \emptyset$\\
		\For{$q\in L \mid q\neq p$}
			{
    			$\lambda_q \leftarrow$ measure$(D_p \cup D_q, j)$ \\
			$\Psi_p \leftarrow \Psi_p \oplus (q, \lambda_q)$ \\
			}
		$\Lambda_j \leftarrow \Lambda_j \oplus (p, \Psi_p)$  \\ 
		}
	$\Lambda \leftarrow \Lambda \oplus (j, \Lambda_j)$\\
	}
\end{algorithm}

\begin{figure}[b]
\centering
\includegraphics[width=0.6\linewidth]{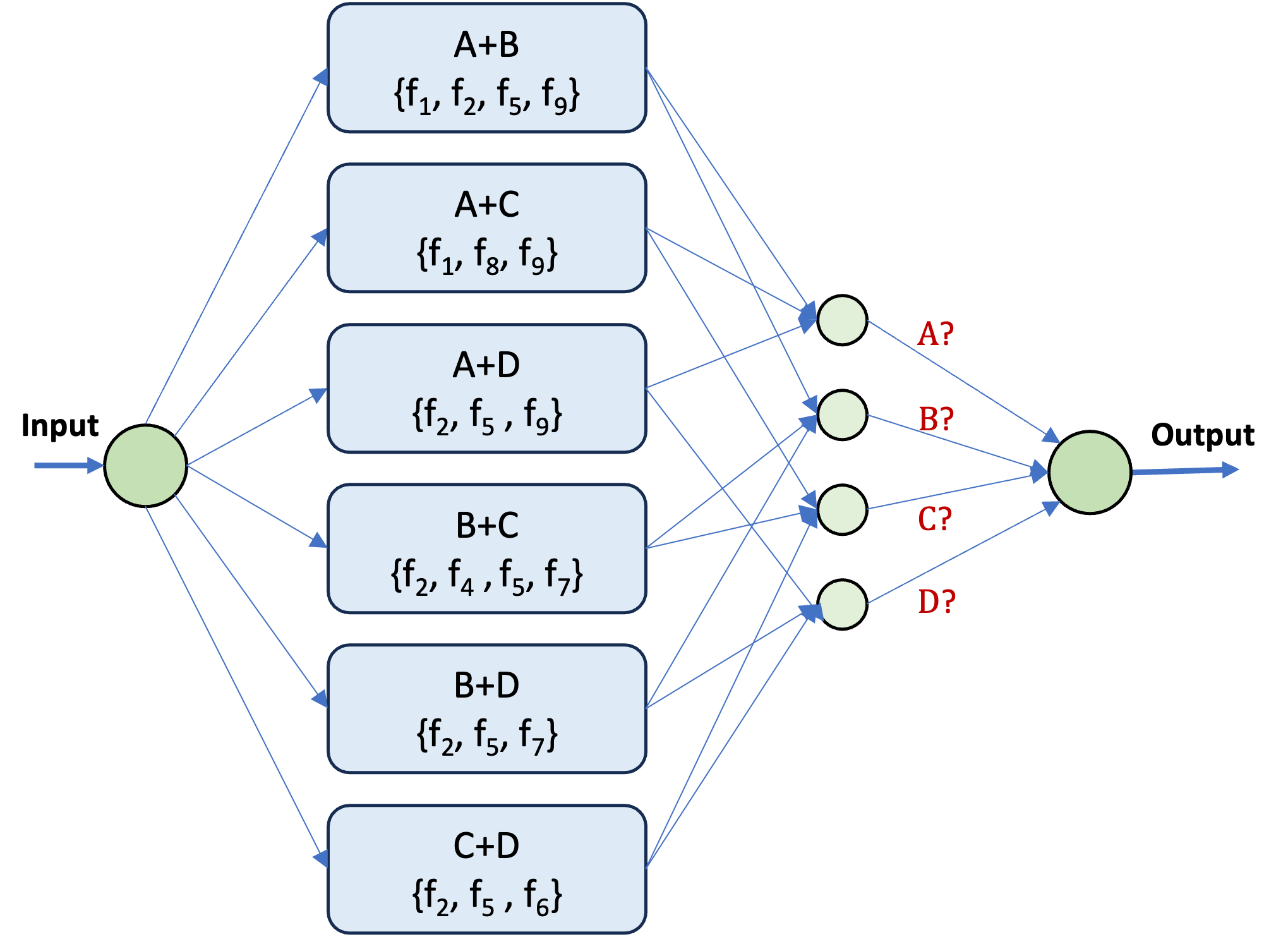}
\caption{Two--layer scheme classification scheme from deep one--versus--each strategy.}
\label{fig:one-against-each_extended}
\end{figure}

This new strategy, which we call \textit{deep one--versus--each} (DOvE), is very interesting because the information contained is richer and, therefore, the classifier ensemble can be more specific in terms of discrimination. Thus, the matrix shown in Table \ref{tab:ex_ranking} could be extended by considering all the paired values before aggregating. From class--pairwise feature information shown in Table \ref{tab:ex_ranking_extended} a more specific classifier ensemble could be designed, as illustrated in Figure \ref{fig:one-against-each_extended}, in which each classifier always discriminates between two classes. In fact, Table \ref{tab:ex_ranking} is an aggregation of Table \ref{tab:ex_ranking_extended}: every row in Table \ref{tab:ex_ranking} (class $p$) is the average of all the rows in Table \ref{tab:ex_ranking_extended} involving the class $p$ ($\forall p \in \mathbb{L}$). It is worth mentioning that features $\{f_4, f_6, f_8\}$ have appeared for the first time in this new classification scheme and were not present earlier.

 \begin{table}[h]
\centering
\caption{Example of result of class--specific strategy including the information of all the pairs of classes.}
\label{tab:ex_ranking_extended}
\begin{tabular}{|c||c|c|c|c|c|c|c|c|c|}
\hline
      & $f_1$ 			& $f_2$ 			& $f_3$ 	& $f_4$ 			& $f_5$ 			& $f_6$ 			& $f_7$ 			& $f_8$ 			& $f_9$ 			\\ \hline \hline
AB & {\color{red}0.9} 	& {\color{red}0.8} 	& 0.1		& 0.2 			& {\color{red}0.6} 	& 0.3				& 0.2 			& 0.3				& {\color{red}0.8}  	\\ \hline
AC & {\color{red}0.8} 	& 0.2 			& 0.2 	& 0.4 			& 0.5 			& 0.4 			& 0.2 			& {\color{red}0.8} 	& {\color{red}0.9}	\\ \hline
AD & 0.4 				& {\color{red}0.8} 	& 0.3 	& 0.3 			& {\color{red}0.7} 	& 0.2				& 0.2 			& 0.1 			& {\color{red}0.7} 	\\ \hline
BC & 0.1 				& {\color{red}0.8} 	& 0.3 	& {\color{red}0.9} 	& {\color{red}0.7} 	& 0.2 			& {\color{red}0.8} 	& 0.1 			& 0.1 			\\ \hline
BD & 0.2 				& {\color{red}0.8} 	& 0.2 	& 0.1 			& {\color{red}0.8} 	& 0.1 			& {\color{red}0.8} 	& 0.2 			& 0.3 			\\ \hline
CD & 0.3 				& {\color{red}0.8} 	& 0.1 	& 0.2 			& {\color{red}0.9} 	& {\color{red}0.9} 	& 0.2 			& 0.3 			& 0.2 			\\ \hline
\end{tabular}
\end{table}

The most powerful aspect of this novel scheme is the introduction of an additional layer. In this setup, each classifier yields an output concerning two distinct classes. Consequently, each node within this new layer analyzes the outputs from $L-1$ classifiers, with each classifier utilizing a unique set of features, to generate a score for a class. This layered approach enriches the classification process by integrating diverse feature sets and classifier insights to arrive at more nuanced class assessments.

\subsection{Class--specific relevance matrix}


The concept of explainability could be enhanced through a more thorough analysis of features from a class--specific perspective. From the information collected in Table \ref{tab:ex_ranking_extended},  we can define a novel matrix, which we call the \textit{class--specific relevance matrix}, and which contains very valuable information on the relevant features among classes. This new matrix significantly contributes to explainability, suggesting that more sophisticated classification schemes could be designed.

When for all pairs including class $p$ the value is greater than the threshold (in red in Table \ref{tab:ex_ranking_extended}), that feature will appear in the class--specific relevance matrix in the diagonal corresponding to class $p$, as shown in Table \ref{tab:relevant_features}. This way, we first calculate all the relevant features in the diagonal: \mbox{$A \rightarrow \{f_9\}$},  \mbox{$B \rightarrow \{f_2, f_5\}$},  \mbox{$C \rightarrow \{\}$}, and  \mbox{$D \rightarrow \{f_2, f_5\}$}. To calculate the feature set for any position in the matrix outside the diagonal, for instance (A,B), we start with the join of diagonal elements of $A$ and $B$, i.e., $\{f_9\}\cup \{f_2, f_5\}$, and then the row $(A,B)$ in Table \ref{tab:ex_ranking_extended} is examined to identify any other feature in red, in this case,  $\{f_1\}$, which is placed in row $(A,B)$ in Table \ref{tab:relevant_features}. For instance, position $(A,D)$ is empty because the join of $A$ and $D$ is exactly what is indicated in row $(A,D)$ in Table \ref{tab:ex_ranking_extended}.



\begin{table}[h]
\centering
\caption{Class--specific relevance matrix.}
\label{tab:relevant_features}
\begin{tabular}{|c |c |c |c |c|} \hline 
		& \hlight{A} 		& \hlight{B} 			& \hlight{C} 			& \hlight{D} 			\\ \hline 
\hlight{A} 	& \ndiag{$\{f_9\}$}  	& \ndiag{$\{f_1\}$} 		& \ndiag{$\{f_1, f_8\}$}  	& \ndiag{$\emptyset$} 	\\ \hline
\hlight{B} 	& --- 				& \ndiag{$\{f_2, f_5\}$} 	& \ndiag{$\{f_4, f_7\}$}  	& \ndiag{$\{f_7\}$}		\\ \hline
\hlight{C} 	& --- 				& --- 					& \ndiag{$\emptyset$}	& \ndiag{$\{f_6\}$}		\\ \hline
\hlight{D} 	& --- 				& --- 					& --- 					& \ndiag{$\{f_2, f_5\}$} 	\\ \hline
\end{tabular} 

\end{table}

\begin{figure}[h]
\centering
\includegraphics[width=0.6\linewidth]{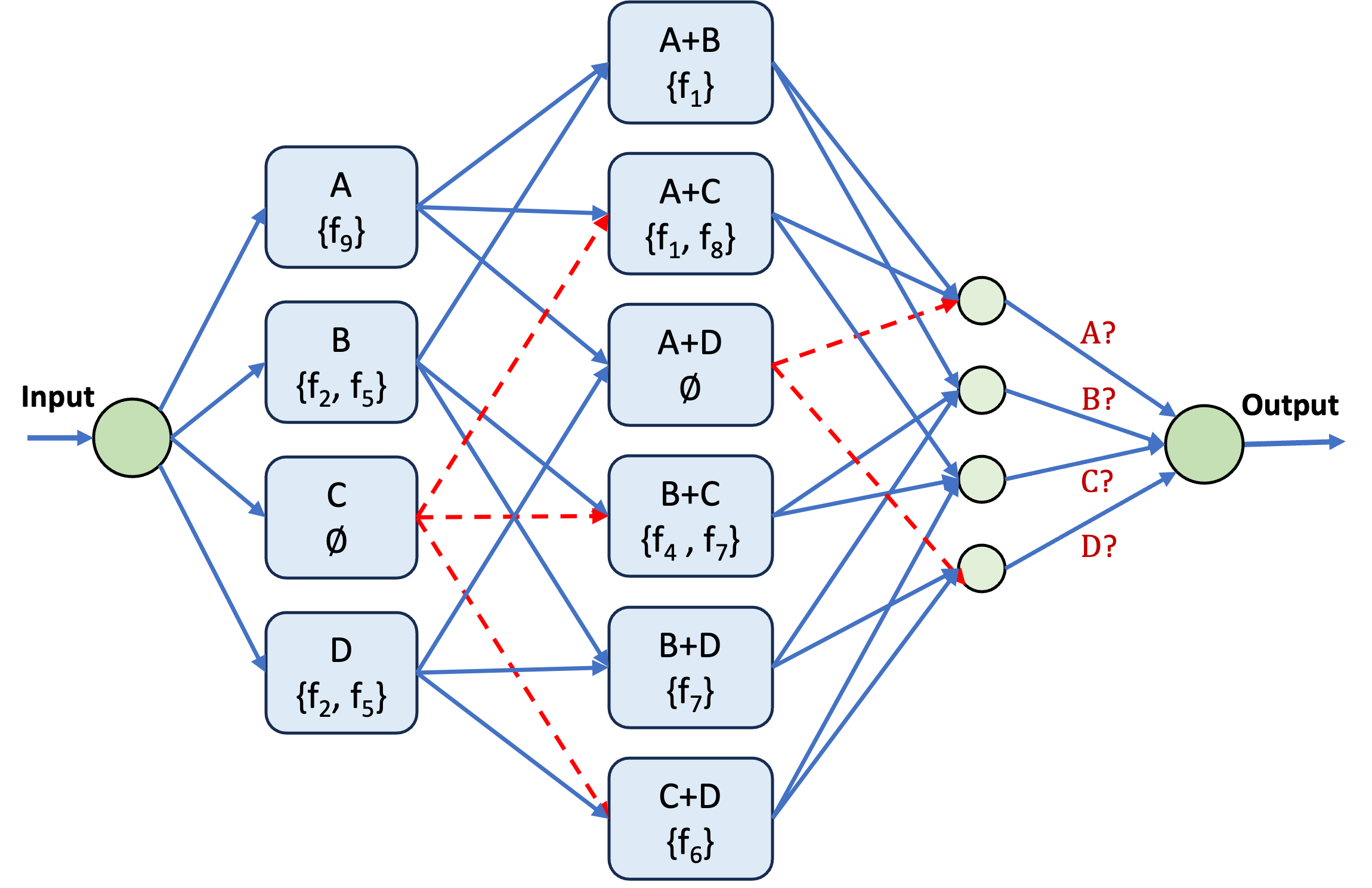}
\caption{Three--layer class--specific classification scheme. (Red dashed lines refer to outputs from empty set of features.)}
\label{fig:one-against-each_extended_deep}
\end{figure}

The classification scheme presented in Figure \ref{fig:one-against-each_extended} can also be inferred from the class--specific relevance matrix shown in Table \ref{tab:relevant_features}. However, from the class--specific relevance matrix it would be possible to further deepen the explainability and design novel classification schemes. For instance, a potential scheme from Table \ref{tab:relevant_features} is introduced in Figure \ref{fig:one-against-each_extended_deep}, which we call \textit{three--layer class--specific classification scheme}. A first layer individually separates classes (taken from the diagonal of class--specific relevance matrix), a second layer examines the discriminatory power of pair of classes, and a third layer integrates all the knowledge for each class. In general, for $L$ classes, the total number of discriminative nodes in the three--layer scheme will be $\binom{L+1}{2}$.

In conclusion, by integrating machine learning explainability with class--specific feature selection, practitioners can obtain a comprehensive and detailed insight into the functioning of models, particularly in situations where distinguishing between multiple classes is paramount. The collaboration of these concepts plays a pivotal role in building machine learning models that are not only accurate but also trustworthy and interpretable.

%
%

\section{Conclusions}

Undoubtedly, the information provided by class--specific techniques is richer than that provided by class--independent techniques. In fact, a class--independent result could be obtained from a class--specific one. The knowledge inherent in the results of class--specific strategies contains a greater level of detail and, therefore, rises to a higher level of explainability. 

Several class--specific approaches mentioned in Related Work have built classifier ensembles based on a one--layer structure. However, more sophisticated decomposable structures can be designed, such as the two--layer or three-layer classification schemes. Therefore, class--specific approaches can substantially contribute to decompose the complexity of classification models, and thus to better understand and interpret the models. Even when using black--box models (e.g., deep learning--based ones) as units of the ensemble, these can use the knowledge from the class--specific relevance matrix to build simpler classifiers, notably reducing the overall complexity of the classification.


The potential for advancement in the class--specific topic is highly promising, given the very sparse scientific literature currently available. Recent research highlights a growing interest in explainability (and its related concepts) within Artificial Intelligence, which is likely poised for significant expansion in the near future. Consequently, developing methods that leverage class--specific strategies, especially in the context of multiclass high--dimensional data, could facilitate knowledge transfer across domains. The scientific landscape, therefore, offers vast and appealing opportunities for future research work.  

\section*{Acknowledgment}

This work was supported by Grant PID2020-117759GB-I00 funded by MCIN/AEI/10.13039/501100011033, the Andalusian Plan for Research, Development and Innovation.

\bibliographystyle{unsrt}  
\bibliography{bibliography}



\end{document}